\pdfoutput=1

\documentclass[11pt]{article}

\usepackage[final]{acl}

\usepackage{times}
\usepackage{latexsym}

\usepackage[T1]{fontenc}

\usepackage[utf8]{inputenc}

\usepackage{microtype}

\usepackage{inconsolata}

\usepackage{amsmath}
\usepackage{times}
\usepackage{latexsym}
\usepackage{booktabs}
\usepackage{microtype}
\usepackage[pdftex]{graphicx}
\usepackage{subfigure}
\usepackage{color,soul} 
\usepackage{xcolor}
\title{Improving Quotation Attribution with Fictional Character Embeddings}

\begin{NoHyper}

\author{Gaspard Michel$^{\dagger\ast}$ \\ \texttt{gmichel@deezer.com} \And Elena V. Epure$^\dagger$  \\ \texttt{eepure@deezer.com}\And
  Romain Hennequin$^\dagger$ \\ \texttt{rhennequin@deezer.com} \AND Christophe Cerisara$^\ast$ \\ \texttt{christophe.cerisara@loria.fr} \AND
  $^\dagger$ \normalfont{Deezer Research, Paris, France} \\ $^\ast$ \normalfont{LORIA, CNRS, Nancy, France}
 }
\end{NoHyper}

\begin{document}
\maketitle
\begin{abstract}
Humans naturally attribute utterances of direct speech to their speaker in literary works.
When attributing quotes, we process contextual information but also access mental representations of characters that we build and revise throughout the narrative.
Recent methods to automatically attribute such utterances have explored simulating human logic with deterministic rules or learning new implicit rules with neural networks when processing contextual information.
However, these systems inherently lack \textit{character} representations, which often leads to errors in more challenging examples of attribution: anaphoric and implicit quotes.
In this work, we propose to augment a popular quotation attribution system, BookNLP, with character embeddings that encode global stylistic information of characters derived from an off-the-shelf stylometric model, Universal Authorship Representation (UAR).
We create DramaCV\footnote{Code and data can be found at \url{https://github.com/deezer/character_embeddings_qa}}, a corpus of English drama plays from the 15th to 20th century that we automatically annotate for Authorship Verification of fictional characters utterances, and release two versions of UAR trained on DramaCV, that are tailored for literary characters analysis.
Then, through an extensive evaluation on 28 novels, we show that combining BookNLP's contextual information with our proposed global character embeddings improves the identification of speakers for anaphoric and implicit quotes, reaching state-of-the-art performance.
\end{abstract}

\section{Introduction}
Authors of literary works employ various devices to create engaging narratives, often combining narration with character dialogues to unveil the plot.
Fictional characters portray themselves through dialogues, revealing aspects of their personality, their own style and ideas about themselves and the fictional world. 
As part of studying characters in digital humanities, automatically identifying utterances and attributing them to characters, also known as quotation attribution, is central \cite{Elson2010, Muzny2017a, Labatut2019, Sims2020}.



Recent approaches to quotation attribution typically succeed at attributing \textit{explicit} quotes but struggle when it comes to \textit{anaphoric} and \textit{implicit} quotes \cite{Muzny2017, CuestaLazaro2022, Vishnubhotla2023, Su2023}.
Explicit utterances occur when the narrator indicates the speaker of a quote with a speech verb and a named mention, while anaphoric quotes are introduced with a speech verb and a pronoun or common noun.
When no narrative information is given about the speaker of the quote, we refer to those as implicit quotes.
Examples of such quotes are given in Figure~\ref{fig:quote_types}.
\begin{figure}
    \small
    \begin{enumerate}
\item[] \textcolor{olive}{"As soon as ever Mr. Bingley comes, my dear,"}  \underline{said Mrs. Bennet,} \textcolor{olive}{"you will wait on him of course."}
\item[] \textcolor{purple}{"No, no. You forced me into visiting him last year, and promised if I went to see him, he should marry one of my daughters..."}
\item[]  His wife represented to him how absolutely necessary such an attention would be from all the neighbouring gentlemen, on his returning to Netherfield.
\item[]  \textcolor{brown}{"'Tis an etiquette I despise,"} \underline{said he.}
    \end{enumerate}
    \caption{Excerpt of \textit{Pride and Prejudice} by Jane Austen (1813). Quotations are colored by quote type: \textcolor{olive}{explict}, \textcolor{purple}{implicit} and \textcolor{brown}{anaphoric}. Speaker information given by the narrator are underlined.}
    \label{fig:quote_types}
\end{figure}

Attributing anaphoric and implicit quotes usually requires fine-grained understanding of contextual information such as discerning discussion patterns and long-range information such as linking pronouns to their canonical entity.
Accessing information of fictional characters (style, persona, gender) also helps disambiguate challenging examples of attribution.
Variations of writing style and long-range contextual dependencies, ubiquitous in fictional works, add an extra layer of complexity, making attribution of non-explicit quotes a complex task to solve even with recent tools. \cite{Sims2020, Vishnubhotla2022}

Some of these shortcomings have been addressed by improving link prediction between pronouns/common nouns and canonical characters \cite{Vishnubhotla2023} or by exploiting better contextual representations \cite{CuestaLazaro2022, Su2023}. Instead, we show that combining fictional character representations encoding global stylistic information and topical preferences with contextual information processed by BookNLP, \footnote{\url{https://github.com/booknlp/booknlp}}, a widely used NLP pipeline designed for literary texts, improves the speaker attribution of anaphoric and implicit quotes.
As explicit utterances are often accurately attributed to their speaker \cite{Vishnubhotla2023}, we assume those quotes given, and use them to derive stylistic information of characters.

Among methods to represent characters from the words they utter, previous works \cite{Li2023, Aggazzotti2023, Michel2024} have explored embeddings encoding topical preferences using SBERT \cite{reimers-gurevych-2019-sentence} or stylistic information using Universal Authorship Representations (UAR) \cite{RiveraSoto2021}, a fine-tuned variant of SBERT trained on the Authorship Verification (AV) task with millions of Reddit users.
UAR's stylometric embeddings built from character utterances were previously shown to distinguish fictional characters among the same work \cite{Michel2024} but considering the discrepancies between Reddit and literary discourse, we question the effectiveness of these embeddings to represent characters.
To overcome this potential limitation, we build variants of UAR tailored for literary texts that we validate on the AV task with drama and novel characters. 

Stylometric character representations computed with our UAR variants significantly improve AV of characters in literary works, while also reaching state-of-the-art quotation attribution accuracy of non-explicit quotes when combined with BookNLP contextual representations. To sum up, our contributions are: 


\begin{itemize}
    \item  We construct DramaCV, a corpus of English drama plays, dating from the 15th to 20th century, focused on the AV of fictional characters within the same work, and train variants of UAR on this dataset, showing that our proposed models significantly improve AV on DramaCV and also generalize better to novels.
    \item Through an extensive evaluation, we show that BookNLP can be easily extended to improve its current performance on the Project Dialogism Novel Corpus (PDNC) \cite{Vishnubhotla2022}, and reaches state-of-the-art attribution accuracy of non-explicit quotes when trained in combination with global character representations computed from their explicit quotes.
\end{itemize}



\section{Related Work}

\paragraph{Quotation Attribution.} \;
Datasets of quotation attribution containing fully annotated English novels are rather scarce, and are often limited to one or two books.
\citet{Elson2010a} introduce the CQSC corpus and attribute automatically extracted quotes to named entities and nominals with a supervised mention ranking model.
Instead, \citet{He2013} attribute quotes directly to \textit{speakers} with a supervised ranking system using features such as speaker alternation patterns and character-level features \cite{He2010}.
Other approaches involve sequence labeling \cite{okeefe-etal-2012-sequence}, Dialogue State Tracking \cite{CuestaLazaro2022} on English novels, or scoring systems on Chinese novels \cite{Chen2021}.
The deterministic sieve-based model of \citet{Muzny2017} regards quotation attribution as a two-step process: quote-\textit{mention} linking and mention-\textit{speaker} linking.
The NLP pipeline dedicated to books, BookNLP, replaced the deterministic sieves with fine-tuned language models.
The link between mention and speaker is carried out by BookNLP's coreference resolution model, and the quote-mention linking is predicted by a BERT model, fine-tuned on LitBank's speaker annotations \cite{Sims2020}.
Recently, \citet{Vishnubhotla2022} introduce the largest-to-date corpus of quotation attribution, PDNC, and show a similar accuracy score of around 63\% for both BookNLP and the sieve-based model.
However, better results were obtained later by fine-tuning BookNLP on PDNC and restricting the predicted coreference chains to resolved characters only \cite{Vishnubhotla2023}.
SIG \cite{Su2023} adapted quotation attribution for encoder-decoder models, showing that predicting both the speaker and the addressees of a quote with BART reaches state-of-the-art accuracy on PDNC.

Interestingly, reported BookNLP attribution results on PDNC often approach perfect accuracy for explicit quotes, indicating that most of the research efforts should be put in the attribution of non-explicit quotes.
Driven by this insight, we exploit explicit quotes uttered by characters to represent them in a latent space, and are the first to show their potential for quotation attribution when combined with contextual representations.

\paragraph{Fictional Character Embeddings.} \;
Understanding fictional characters objectives, personas and style often comes naturally when reading novels.
When accessing information from the fictional world and interactions between characters provided throughout the narrative, we infer mental representations of characters that are revised when accessing new information \cite{Gernsbacher1992, Culpeper1996}.
Previous computational works on this aspect of fictional character understanding have explored representing fictional characters in a latent space, often targeting a specific aspect, such as persona, style or descriptive elements.
These computational representations of characters are often used for further analysis of large literary corpora, trying to understand character distinctiveness between authors \cite{Bamman2014} or narrative comprehension \cite{Brahman2021}.
Other works focus on using fictional characters embedings to improve character-level tasks such as coreference resolution and character linking \cite{Li2023} or contextual and book-level tasks such as narrative question answering \cite{Inoue2022}.

Here, we propose to represent character from \textit{what they say} and \textit{how they say it}, building literary embeddings encoding style and topical preferences of characters, derived from models trained on the Authorship Verification (AV) of fictional characters within the same work.
Most works exploring the AV of fictional characters focus on drama characters because of the availability of large annotated corpora and aim at capturing syntactic, lexical and phonological variations that occur in their direct speech using statistical measures of distinctiveness \cite{Dinu2017, Vishnubhotla2019, Sela2023}.


With the recent introduction of PDNC \cite{Vishnubhotla2022}, a corpus of 22 novels annotated with quotation attribution\footnote{6 additional novels were released a year later, totaling 28 annotated novels.}, analysis on fictional characters in novels can be done at a larger scale. 
\citet{Michel2024} provide a comparative study of PDNC's characters, converting the original PDNC quotation attribution task to the AV of fictional characters and showing that UAR \cite{RiveraSoto2021} embeddings derived from a small collection of utterances are able to distinguish characters within the same work.
However, Reddit -- the target domain of UAR -- is inherently different from traditional novels that use conventions on grammar, punctuation and format, motivating the need for a variant of UAR tailed for literary texts.  

\begin{table*}[h!]
    \centering
    \small
    \begin{tabular}{l|c|cc|cc}
    \toprule
    & \textbf{Split} & \textbf{Segments} & \textbf{Utterances} &\textbf{Queries} & \textbf{Targets/Query (avg)}\\
    \midrule
    & Train & 1507 & 263270 & 5392 & 5.0\\ 
    \textbf{Scene} & Val & 240 & 50670 & 1557 & 8.8\\ 
    & Test & 203 & 41830 & 1319 & 8.7\\
    \midrule
    & Train & 226 & 449407 & 4109 & 90.7\\
    \textbf{Play} & Val & 30 & 63934 & 917 & 55.1 \\ 
    & Test & 31 & 74738 & 1214 & 108.5\\
    \midrule
    \midrule
    \textbf{PDNC} & Explicit & - & 6303 & 562 & 11.2\\
    \bottomrule
    \end{tabular}
    \caption{Summary statistics of the Scene and Play instances of DramaCV by split. A segment in the Scene split is a single scene or act in a play, while a segment in the Play split is the entire play. For each segment, the targets are collections of utterances uttered by all participating characters in the segment. The Explicit split of PDNC \cite{Michel2024} that we use to test generalization is also displayed for comparison.}
    \label{tab:dramachar_stat}
\end{table*}

\section{Representing Fictional Characters}

In this section, we present our approach to build and evaluate character representations tailored for literary texts, focusing on aspects of style and topical preferences of characters.

To build these representations, we train variants of UAR from scratch on the Authorship Verification (AV) of characters in the literary domain, considering a fictional character as an \textit{author} in the AV terminology.
The AV of fictional characters aims at answering the following problem: given two collections of quotes that have been uttered by characters within the same fictional work, predict whether they were spoken by the
same character or not.
Although closely linked to Authorship Attribution, we note that this framework is different for the following reasons: 1) we do not have access to a list of canonical characters for every literary work and 2) we compare characters within the same story, implying that each quote is inherently influenced by the author's style.

UAR adapts SBERT by directly encoding collections of documents, and is trained to distinguish if two collections of documents have been written by the same author or not.
Although the Reddit version of UAR appears to perform well on PDNC \cite{Michel2024}, which is on-par with its zero-shot domain transfer abilities, we question its suitability for literary analysis and propose to train instead a domain-specific variant of UAR tailored for literary texts.

Only a few datasets containing annotations of the speaker of an utterance of direct speech in full novels exist, as annotating entire works is costly.
Instead, prior works have considered using movie \cite{Azab2019, Sang2022, Li2023} or drama scripts \cite{Vishnubhotla2019, Fischer2019} as they come naturally annotated with the speaker of each utterance.
We follow this approach and propose DramaCV (Drama Character Verification), a dataset of English drama plays focused on the AV of fictional characters.
We chose to use drama plays as they are mostly similar to novels in terms of writing conventions, and because we were able to extract a number of characters and utterances an order of magnitude above the available datasets of annotated novels. 
The rest of this section evaluates our variants of UAR trained on DramaCV on the test set of DramaCV and on fictional characters participating in the 28 PDNC novels.





\subsection{DramaCV}

We extract 499 plays dating from the 15th to 20th century from Project Gutenberg\footnote{\url{https://gutenberg.org/}} with the GutenTag software \cite{brooke-etal-2015-gutentag} that we automatically parsed to create a character list and to attribute each line to its speaker.
Then, we construct two instances of DramaCV:

\paragraph{Scene.} We split each play in scenes, a small segment unit of drama that is supposed to contain actions occurring at a specific time and place with the same characters.
If a play has no {\small{\texttt{<scene>}}} tag, we instead split it in acts, with the {\small{\texttt{<act>}}} tag.
Acts are larger segment units, composed of multiple scenes.
For this split, we only consider plays that have at least one of these tags.
Although scenes usually contain a few characters, they represent a challenge for CV models as participating characters are akin to discuss a similar topic.
Thus, stylistic cues of characters are likely to be useful to distinguish characters within the same scene.

\paragraph{Play.} We do not segment play and use all character lines in a play.
Compared to the scene segment, the number of candidate characters is higher, and discussions could include various topics. 
Hence, topical preferences of characters might be an additional useful feature to distinguish characters.

\phantom{text}

After parsing, a total of 169 plays for the Scene split, and 287 plays for the Play split remained.
For both splits, we use 80\% of the plays for training, 10\% for validation and 10\% for test.
Following the AV setup, we build queries and targets using collections of character utterances.
In this setup, the objective is to recognize if a query and a target have been uttered by the same character or not. 
We construct queries in the validation and test splits by randomly sampling half of the utterances of a character, 
and we use the remaining half as the target.
The train splits only consider characters that uttered at least 16 lines, such that we can build queries and targets with at least 8 utterances.
More details on the construction of queries and targets can be found in Appendix~\ref{sec:appendix_a}.
Summary statistics for both instances of DramaCV are presented in Table~\ref{tab:dramachar_stat}.


\subsection{Literary UAR}
\label{sec:char_repr}
We train UAR from scratch using its public implementation on DramaCV.
For both instances, we use the pre-trained \texttt{all-distilroberta-v1} as the base encoder, and employ a learning rate of 2e-5 for 20 epochs.
We set the dimension of the model to $d=512$ and use a maximum sequence length of 64 tokens for all utterances. 
During training, we build queries and targets by randomly sampling 8 utterances to represent characters.
A batch size of 1 play is used when training on the Play split, and 8 scenes are used when training on the Scene split.
During inference, we do not restrict the number of utterances in a collection to reflect situations where we have access to only less than 8 utterances, or to more than 8 utterances.  

A supervised contrastive objective \cite{Khosla2020} is used where in-batch negatives considered are characters within the same segment unit.
We motivate this choice for each dataset instance:

\noindent \textbf{Scene.} \;In a scene, characters are likely to discuss similar topics. By restricting the objective to characters within a scene, we force the model to rely on stylistic cues to distinguish lines of characters.

\noindent \textbf{Play.}  We want to avoid capturing the writing style of authors to distinguish between collections of character lines. Thus, we restrict the set of in-batch negatives to characters within the same work.

We compare our variants of UAR, $\text{UAR}_{Scene}$\footnote{\url{https://huggingface.co/gasmichel/UAR_Scene}} and $\text{UAR}_{Play}$\footnote{\url{https://huggingface.co/gasmichel/UAR_Play}} against their Reddit counterpart, $\text{UAR}_{Reddit}$ and SBERT.
Following previous work \cite{Aggazzotti2023, Michel2024}, we create SBERT character embeddings by encoding independently each utterance in a feature $d$-dimensional vector\footnote{We use \texttt{all-mpnet-base-v2}.}, followed by a coordinate-wise mean of all utterance vectors. 
Instead, UAR directly encodes a collection of utterances into a $d$-dimensional vector.

Given a collection of a character utterances (query) and collections of all characters occurring in the same segment (targets, that include a target uttered by the same character as the query), we evaluate the ability of these models to yield a higher cosine-similarity to the target uttered by the same character than to targets uttered by other characters using Area Under the Receiver Operating Characteristic curve (AUC).
We use this evaluation setup on fictional characters for both test-sets of DramaCV and for the 28 novels of PDNC on the Explicit split of \citet{Michel2024}.
More details on the evaluation can be found in Appendix~\ref{sec:appendix_a}.


\begin{table}[t!]
    \centering
    \small
    \begin{tabular}{l|cc|cc}
    \toprule
    & \multicolumn{2}{c}{\small{\textbf{DramaCV}}} & \multicolumn{2}{c}{\small{\textbf{PDNC - Exp}}} \\
    \midrule
    & \small{\textbf{Scenes}} & \small{\textbf{Plays}} & \small{\textbf{CC}} & \small{\textbf{CQ}}\\
    \midrule
    SBERT & 72.0 \scriptsize{(5)} & 78.8 \scriptsize{(10)} & 63.8 \scriptsize{(16)} & 54.4 \scriptsize{(5)}\\
    $\text{UAR}_{Reddit}$ & 68.6 \scriptsize{(7)} & 74.5 \scriptsize{(11)} & 81.1 \scriptsize{(9)} & 54.1 \scriptsize{(4)}\\
    \midrule
    $\text{UAR}_{Scene}$ & \textbf{82.3} \scriptsize{(6)} & - & 81.3 \scriptsize{(10)} & \textbf{58.9} \scriptsize{(4)}\\
    $\text{UAR}_{Play}$ & - & \textbf{84.2} \scriptsize{(9)} & \textbf{86.5} \scriptsize{(8)} & 58.5 \scriptsize{(5)}\\

    \bottomrule
    \end{tabular}
    \caption{Average AUC (\%) on test plays in DramaCV and on novels in the Explicit split of PDNC with standard deviations between parentheses.}
    \label{tab:av_results}
\end{table}

\subsection{Authorship Verification Results}
We report AUC for both test splits of DramaCV and the Explicit split of PDNC in Table~\ref{tab:av_results}.
Interestingly, $\text{UAR}_{Reddit}$ performs worse on both splits compared to SBERT and in-domain UARs.
$\text{UAR}_{Scene}$ increases AUC on its training domain up to 10 points compared to SBERT, indicating that the model is able to capture stylistic cues specific to drama that help at distinguishing utterances of characters discussing similar topics.
The increase of $\text{UAR}_{Play}$ is slightly lower, but the high AUC indicates that it succeeds at distinguishing characters within the same play.
The high standard deviation across plays suggests that task difficulty is not distributed uniformly among these literary works.
Among best-performing plays, we found works by canonical authors such as Shakespeare's \textit{Hamlet}, Shaw's \textit{John Bull's Other Island} or Ben Jonson's \textit{The Alchemist}
(see complete list in Appendix~\ref{sec:appendix_b}).

Looking at generalization results of PDNC, $\text{UAR}_{Reddit}$ performs better than SBERT, and almost equals $\text{UAR}_{Scene}$ in the Character-Character setup (CC), which compares similarity of character representations.
We find significant improvement with $\text{UAR}_{Play}$ over the baselines, suggesting that training UAR on drama allows better domain adaptation to literary novels.
We also remark that this adaptation is data-efficient, as training $\text{UAR}_{Scene}$  with around 5000 drama characters is enough to match the performance of $\text{UAR}_{Reddit}$ that has been trained on millions of different users.

The advantages of domain adaptation can also be seen in the Character-Quote (CQ) evaluation setup that aims at attributing utterances to their speaker by comparing the similarity between character and quote embeddings computed with the same model.
Here, the two variants of UAR trained on drama show a significant increase in AUC over the baselines, indicating that character representations built with our proposed UARs are more similar to embeddings of their other quotes.
However, AUC scores for the CQ experiment are low, showing that the task of attributing quotes requires more than only comparing similarities of character and quote embeddings.

\begin{figure*}
    \centering
    \includegraphics[width=\linewidth]{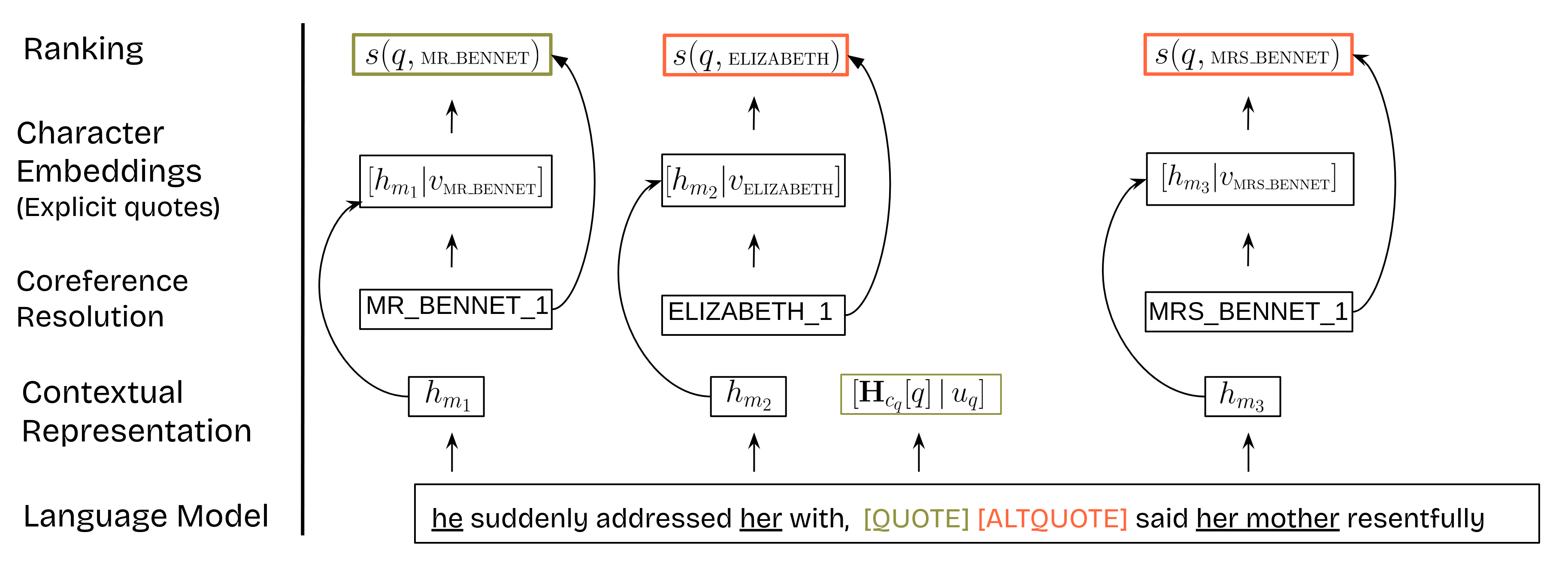}
    \caption{Detailed process of BookNLP's quotation attribution system. We inject extra-information in the form of character embeddings computed from explicit quotes $v_{\text{entity}(m)}$ in the fourth step, and quote embeddings $u_q$ in the second step.}
    \label{fig:book_nlp}
\end{figure*}

\section{Quotation attribution}

Motivated by the ability of UAR to encode character utterances in a representation space that contains useful, though insufficient, information for quotation attribution, we evaluate in the next section the impact of combining these global character embeddings with BookNLP's contextual information.
BookNLP is a Natural Language Processing pipeline dedicated to English books, popular among practitioners of Digital Humanities.

\subsection{BookNLP}
\label{sec:booknlp}
BookNLP's pipeline includes, among others, Part-of-speech tagging, Named Entity Recognition, Coreference Resolution and quotation attribution.
All components of the pipeline have been trained on LitBank \cite{bamman-etal-2019-annotated, bamman-etal-2020-annotated, Sims2020}, a corpus containing 100 chapters from public domain books. Since its quotation attribution model is not part of a publication, we will review its details in the following paragraphs.

BookNLP carries out quotation attribution in four independent steps: 1) tokenization and named entity recognition, 2) character name clustering, 3) pronominal coreference resolution, and 4) quote-mention linking. 
The first three steps are dedicated to extract named, nominal and pronominal mentions along with the predicted character they refer to.
Using extracted mentions in the surrounding context of a quote, the final step, quote-to-mention linking, aims at predicting which mention is the one that refers to the true speaker of the quote.
In other words, it finds the information given by the narrator about who is speaking among the narrative parts in the vicinity of quotes.  
It is achieved by a fine-tuned BERT model \cite{devlin-etal-2019-bert}.
We give further details of the quote-mention linking step below.





\begin{table*}[t!]
    \centering
    \small
    \begin{tabular}{l|c|cc|cc}
    \toprule
    &  \textbf{Overall} & \textbf{Non-Explicit} & \textbf{Explicit} & \textbf{Anaphoric} & \textbf{Implicit}\\
    \midrule
    SIG &  $72^+$  & $70.0^+$ & - & - & -\\
    ChatGPT & $71^+$ & $70.0^+$ & - & -\\
    \midrule
    \midrule
    Unanswerable (\%)  & 14.1 \scriptsize{(4.4)} & 20.9 \scriptsize{(6.7)} & 0 & 24.5 \scriptsize{(8.1)} & 19.8 \scriptsize{(8.6)}\\
    \midrule
    $\text{BookNLP+}_{reimp}$ & $78.5^{\phantom{\star\star}}$ \scriptsize{(4.0)} & $68.9^{\phantom{\star\star}}$ \scriptsize{(4.4)} & 98.6 \scriptsize{(1.2)} & $70.2^{\phantom{\star}}$ \scriptsize{(7.0)} & $66.4^{\phantom{\star}}$ \scriptsize{(5.7)}\\
    \midrule
    \;\; + SBERT & $79.2^{\star\star}$ \scriptsize{(3.8)} & $70.0^{\star\phantom{\star}}$ \scriptsize{(4.5)} & 98.3 \scriptsize{(1.1)} & $70.8^{\phantom{\star\star}}$ \scriptsize{(7.3)} & $68.1^{\star}$ \scriptsize{(5.5)}\\
    \;\; + $\text{UAR}_{Reddit}$ & $80.1^{\star\star}$ \scriptsize{(3.9)} & $71.1^{\star\star}$ \scriptsize{(4.7)} & 98.7 \scriptsize{(1.1)} & $71.5^{\phantom{\star\star}}$ \scriptsize{(7.9)} & $\mathbf{69.6}^{\star}$ \scriptsize{(5.3)}\\
    \;\; + $\text{UAR}_{Play}$ & $80.0^{\star\phantom{\star}}$ \scriptsize{(3.8)} & $70.9^{\star\phantom{\star}}$ \scriptsize{(4.4)} & 98.7 \scriptsize{(1.0)} & $71.1^{\phantom{\star\star}}$ \scriptsize{(7.9)} & $69.5^{\star}$ \scriptsize{(5.7)}\\
    \;\; + $\text{UAR}_{Scene}$ & $\mathbf{80.2}^{\star\phantom{\star}}$ \scriptsize{(3.8)} & $\mathbf{71.2}^{\star\phantom{\star}}$ \scriptsize{(4.3)} & 98.7 \scriptsize{(1.0)} & $\mathbf{71.7}^{\star\star}$ \scriptsize{(8.1)} & $\mathbf{69.6}^{\star}$ \scriptsize{(5.4)}\\
    \bottomrule
    \end{tabular}
    \caption{Cross validation accuracy (\%) of quotation attribution on the PDNC dataset. Standard deviations in parentheses are calculated across splits. We take the reported accuracy of SIG and ChatGPT (accuracies for Implicit and Anaphoric quotes are not reported ) from \cite{Su2023} (+). Statistical significance against $\text{BookNLP+}_{reimp}$ from paired t-test  is denoted by $^{\star}$ (5\%) and $^{\star\star}$ (10\%).}
    \label{tab:main_res}
\end{table*}

Let $D = (t_1,\dots t_n)$ be a tokenized document and $q = (t_i, \dots, t_{j})$ be a character utterance in this document, starting at token $i$ and ending at token $j$.
A contextual segment is computed for each quote, $ c_q = [\, c_{q}^{left} \; | \, q \, | \; c_{q}^{right} \,]$, where $c_{q}^{left}$ and $c_{q}^{right}$ are left and right contextual information of length $w$ tokens and $[\, | \,]$ denotes concatenation.
Additionally, the full quote $q$ is replaced by a special token \texttt{[QUOTE]} and all other quotes occurring in the contextual segment are replaced by a special token \texttt{[ALTQUOTE]}, such that the contextual segment mainly reflect narrative parts.
Each segment $c_q$, containing masked quotes and narrative parts, is fed to BERT, producing a contextual representation $\mathbf{H}_{c_q}$. Then an unary compatibility score between the (masked) quote token $q$ and a mention-span $m = (t_{m_s}, \dots, t_{m_e})$ occurring in the segment is calculated as follows: 
\begin{align}
    h_m &= \frac{1}{|m|}\sum\limits_{i=m_s}^{m_e} \mathbf{H}_{c_q}[i]\\
\label{eq:score}
    s(q,m) &= \phi([ \mathbf{H}_{c_q}[q] \; | \; h_m])
\end{align}

\noindent where $\mathbf{H}_{c_q}[i]$ is the BERT representation of token $i$ while $|m|$ is the token length of mention $m$ and $\phi$ is a feed-forward neural network with one hidden layer and a tanh activation. Given a set of candidates $Y(q) = (m_1, \dots, m_k)$, the model is trained to optimize the log-likelihood of mentions referring to the true speaker of $q$. 
Figure~\ref{fig:book_nlp} summarizes this process.



The default window size is $w=50$ tokens, and the hidden layer dimension is $100$.
During training and inference, only a maximum of $k=10$ mention candidates are considered. 

In our experiments, we modify these default parameters by increasing the narrative window size to $w=100$ tokens, and using a relu activation and a hidden dimension of $d=512$ in the feed-forward neural network.
We also replace BERT by SpanBERT-large \cite{Joshi2020}, computing mention representations by concatenating SpanBERT embeddings of the first and last tokens.
At inference, we do not truncate the number of candidates and instead use all candidate mentions available in the context window.

\subsection{Integrating Character Embeddings}

Let $E(c) = (q_i, \dots, q_j)$ be the collection of all explicit quotes uttered by character $c$.
We embed this collection of quotes using one of the model described in Section~\ref{sec:char_repr}, yielding a vector representation $v_c$ for each character.
If a character has not been explicitly quoted, we set its vector representation to zeros.
Additionally, we embed each target quote $q$ independently with the same model to get a vector representation $u_q$. We modify Equation~\ref{eq:score} to combine BookNLP's contextual representation with these character and quote representations: 
\begin{equation}
\label{eq:updated_score}
    s(q,m) = \phi([ \mathbf{H}_{c_q}[q] \; | \; h_m \; | \; v_{\text{entity}(m)} \; | \; u_q])
\end{equation}
where $\text{entity}(m)$ is the character predicted by the coreference resolution step for mention $m$.

Equation~\ref{eq:updated_score} allows BookNLP to rely on external information by injecting global character information and compatibility between quote $q$ and candidate character $c$.

\subsection{Experimental Setup}

Following \citet{Su2023}, we use the first version of PDNC, which contains 22 annotated novels. 
We focus on \textit{major} and \textit{intermediate} characters,
who are characters that uttered at least 10 quotes in the novel.
This allows to avoid the long tail of minor characters and to compare our results to the current state-of-the-art, SIG and ChatGPT (\textit{gpt-3.5-turbo-0613}).

We base our implementation on BookNLP+ \cite{Vishnubhotla2023}, that restricts coreference predictions to chains that can be successfully resolved to a character entity from the annotated PDNC character lists.
We follow the same evaluation protocol, using their provided cross-validation split at the novel level, reporting the average accuracy over each split.   

We train our implementation of BookNLP+, $\text{BookNLP+}_{reimp}$ with the modifications mentioned in Section~\ref{sec:booknlp}, and add character embeddings encoded by different models using Equation~\ref{eq:updated_score}.
We use AdamW optimizer with a learning rate of 5e-6 and train each model for 20 epochs.

\subsection{Results}

Our main results are displayed in Table~\ref{tab:main_res}.
Our implementation of BookNLP+ reaches 68.9\% accuracy on non-explicit quotes, compared to the 53\% reported in \citet{Vishnubhotla2023}.
This large increase is mostly due to the replacement of BERT-base with SpanBERT-large.
Over all quotes, $\text{BookNLP+}_{reimp}$ largely outperforms SIG and ChatGPT.

Combining BookNLP+ with SBERT character embeddings  matches the accuracy of SIG and ChatGPT on non-explicit quotes.
The gains are larger for all variants of UAR, reaching 71.2\% accuracy on non-explicit quotes with our proposed $\text{UAR}_{Scene}$. 
As expected, we found that these gains are largely due to the improved attribution performance of implicit quotes, increasing the accuracy by more than 3 points with all UAR models.

\begin{figure}[t!]
    \centering
    \includegraphics[width=0.81\linewidth]{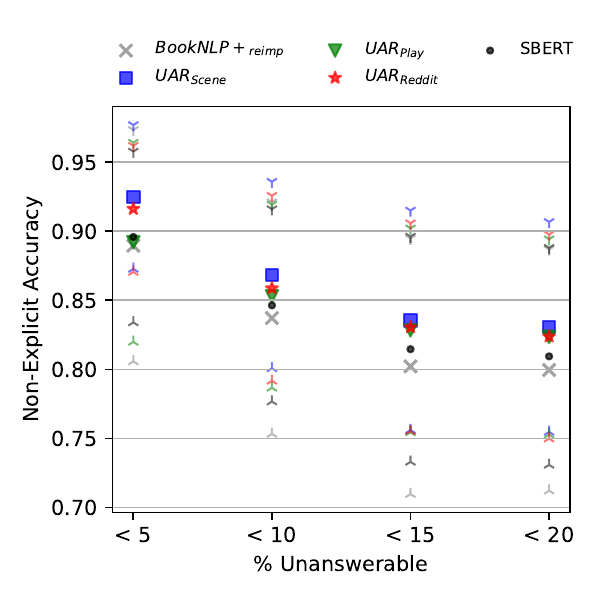}
    \caption{Accuracy results on non-explicit quotes averaged over subsets of novels that have a low percentage of unanswerable quotes, with standard deviations.}
    \label{fig:res_per_error}
\end{figure}

Interestingly, although our proposed variants of UAR tailored for literary texts are better at the Authorship Verification of fictional characters, we only see a small -- though statistically significant -- increase between $\text{UAR}_{Scene}$ and its Reddit counterpart on non-explicit attribution accuracy.
The reasons could be twofold.
First, BookNLP's system only focuses on finding the right narrative information that helps identifying the speaker of a quote.
However, our embeddings mostly reveal stylistic cues of characters, which are not encoded by BookNLP.
Therefore, our implementation provides external character information that might not align perfectly with BookNLP's contextual decisions, limiting their potential combination.
Second, we add character information based on noisy coreference chains, which might hinder the model's performance.
We believe further research should be conducted into finding better combinations between character embeddings and BookNLP.

The high standard deviations across splits reveals that the task is harder for some novels than others.
Looking at the percentage of unanswerable quotes (quotes that do not have a candidate mention referring to their speaker in their context window) provides a reason for such high variations: it ranges from 0 to 56\% per novel.
Figure~\ref{fig:res_per_error} shows the attribution accuracy of non-explicit quotes when only considering novels that have a low percentage of unanswerable quotes.
Only 4 novels have a percentage lower than 5\%, but 17 out of the 22 have a percentage lower than 20\%.
As expected, we see higher performance for all models, and the gain from UAR character embeddings is more clear.

\begin{table}[t!]
    \centering
    \small
    \begin{tabular}{l|ccc}
    \toprule
    & $\Delta$ \textbf{Non-Exp} &  $\Delta$ \textbf{Ana} &  $\Delta$ \textbf{Imp}\\
    \midrule
    SBERT & $-0.0\phantom{^{\star\star}}$ & $-0.0\phantom{^{\star}}$ & $-0.1\phantom{^{\star}}$  \\
    $\text{UAR}_{Reddit}$ & $-0.1\phantom{^{\star\star}}$ & $-0.0\phantom{^\star}$ & $-0.2\phantom{^{\star}}$\\
    \midrule
    $\text{UAR}_{Play}$ & $-0.2^{\star\star}$ & $-0.1\phantom{^{\star}}$ & $-0.3^\star$\\
    $\text{UAR}_{Scene}$ & $+0.1\phantom{^{\star\star}}$ & $+0.1\phantom{^{\star}}$ & $+0.1\phantom{^{\star}}$\\
    \bottomrule
    \end{tabular}
    \caption{Original accuracy (Table~\ref{tab:main_res}) variation when replacing gold explicit attributions with attributions predicted by $\text{BookNLP+}_{reimp}$. Statistical significance against original accuracy from paired t-test is indicated with  $^\star (5\%)$ and $^{\star\star} (10\%)$ }
    \label{tab:variation_res}
\end{table}

Let us note that $\text{UAR}_{Scene}$ is consistently better than $\text{UAR}_{Play}$, though the latter is better at distinguishing PDNC characters in the CC experiment.
Restricting UAR to encode stylistic information by using collections of quotes with similar topic seems to generalize better to quotation attribution.

To construct character embeddings, we relied on gold-labeled attributions of explicit quotes.
We report in Table~\ref{tab:variation_res} the accuracy variation when replacing the gold attributions with speaker predictions from $\text{BookNLP+}_{reimp}$.
Only $\text{UAR}_{Play}$ shows a statistically significant decrease in accuracy, although this variation remains very small.
This validates our intuition that explicit quotes can be extracted in a first step and that the eventual noise contained in the resulting character embeddings does not harm their performance for quotation attribution.

Finally, we present quotation attribution results for 6 annotated novels from the second release of PDNC in Table~\ref{tab:third_res}.
We use the model trained on the first cross-validation split to predict attributions in each of these novels.
The impact of using literary character embeddings is more clear: both of our proposed variants of UAR show increased attribution accuracy against their Reddit counterpart.
Besides, the increase in accuracy over non-explicit quotes is again mostly attributed to an increase in implicit speaker identification.

\section{Conclusion}

In this work, we evaluated fictional character embeddings constructed with the explicit quotes they utter, showing that they improve quotation attribution accuracy of non-explicit quotes on PDNC when used in combination with BookNLP's contextual representations.
We created DramaCV, a corpus of English drama plays focused on the Authorship Verification of fictional characters and trained two variants of UAR on this dataset.
Our proposed models show better performance when distinguishing fictional characters in both DramaCV and PDNC.
Character embeddings derived from these models are further used for quotation attribution on PDNC, reaching state-of-the-art accuracy when combined with BookNLP+.
The impact of using literary character embeddings against embeddings derived from a model trained on Reddit is small, though statistically significant, suggesting superiority of in-domain character representations for quotation attribution.
    
We proposed a straightforward way to combine fictional character embeddings with contextual representations, using BookNLP as an example.
Our approach can be adapted to other models: cross-attention over character embeddings could be used for encoder-decoder models, while modality alignment \cite{zhang-etal-2023-speechgpt} is a solution for decoder-only models.
We leave potential improvements to the combination of fictional character embeddings and contextual representations to future work.  

\begin{table}[t!]
    \centering
    \small
    \begin{tabular}{l|ccc}
    \toprule
    &\textbf{Non-Exp} & \textbf{Ana} & \textbf{Imp} \\
    \midrule
    \footnotesize{Unanswerable (\%)} & 20 & 21 & 18\\
    $\text{BookNLP+}_{reimp}$ & $69.6^{\phantom{\star\star}}$ & $72.3^{\phantom{\star\star}}$ & $67.6^{\phantom{\star}}$  \\
    \midrule
    \;\; + SBERT & $69.3^{\phantom{\star\star}}$  & $71.6^{\phantom{\star\star}}$ & $68.0^{\phantom{\star}}$ \\
    \;\; + $\text{UAR}_{Reddit}$ & $70.5^{\star\star}$  & $72.4^{\phantom{\star\star}}$ &
    $69.4^{\star}$\\
    \midrule
    \;\; + $\text{UAR}_{Play}$ & $71.1^{\star\phantom{\star}}$& $72.8^{\star\star}$ & $\mathbf{70.1}^{\star}$  \\
    \;\; + $\text{UAR}_{Scene}$ & $\mathbf{71.2}^{\star\phantom{\star}}$& $\mathbf{73.1}^{\phantom{\star\star}}$ & $70.0^{\star}$ \\
    \bottomrule
    \end{tabular}
    \caption{Average quotation attribution accuracy (\%) over the 6 additional PDNC novels. Statistical significance from paired t-test against $\text{BookNLP+}_{reimp}$ is denoted by $^{\star}$ (5\%) and $^{\star\star}$ (10\%).}
    \label{tab:third_res}
\end{table}

We found that BookNLP's attribution accuracy is not distributed uniformly across novels, mainly caused by noisy coreference resolution, resulting in many quotes not having contextual candidate mentions that refer to their true speaker.
We believe that including our character embeddings in the coreference resolution model as in \citet{Li2023} might also improve its performance and thus increase overall attribution accuracy.
Another potential solution could include unmasking the closest quotes in the vicinity of a target quote that are likely to contain addressee information.
We leave better combination of fictional character embeddings with BookNLP for future exploration.

\section{Limitations}

In this work, we focused on integrating fictional character embeddings in encoder-only models.
We have not tested with other types of models such as encoder-decoder and decoder only, but we believe that integrating our embeddings in those frameworks is possible and is likely to improve their performance on the quotation attribution task.
Our first set of experiments corroborate this assumption, on the usefulness of our proposed character embeddings in distinguishing fictional characters.

We also proposed models that can be used to create character representations.
These models are trained on a corpus of drama plays, which spans almost 400 years of work.
It is unclear how the variation of writing styles that have occurred throughout these centuries affects the information contained in these embeddings, and how they can generalize to new literary styles.
This remark also extends to our results on quotation attribution.
We presented results on a corpus containing novels from the 19th to early 20th century ; which is closer in terms of writing style to DramaCV than newer novels published in the 21th century.
It is thus unclear how our results generalize to these newer novels, but we strongly believe that accurate character embeddings improve quotation attribution regardless of the writing style of the author.

We only focused on representations encoding stylistic cues and topical preferences and showed that they improve quotation attribution.
However, this type of character embedding might be incomplete and other aspects of characters can be used in addition to our proposed embeddings (persona, gender, objective, or social status).
This might allow to build accurate representations of characters that are likely to improve even more quotation attribution accuracy.

\section{Acknowledgements}

This work was partly performed using HPC resources from GENCI-IDRIS and Grid500.

\bibliography{custom, bilbio, anthology}

\begin{thebibliography}{37}
\providecommand{\natexlab}[1]{#1}

\bibitem[{Aggazzotti et~al.(2023)Aggazzotti, Andrews, and Smith}]{Aggazzotti2023}
Cristina Aggazzotti, Nicholas Andrews, and Elizabeth~Allyn Smith. 2023.
\newblock \href {https://arxiv.org/abs/2311.07564} {Can authorship attribution models distinguish speakers in speech transcripts?}
\newblock \emph{Preprint}, arXiv:2311.07564.

\bibitem[{Bamman et~al.(2020)Bamman, Lewke, and Mansoor}]{bamman-etal-2020-annotated}
David Bamman, Olivia Lewke, and Anya Mansoor. 2020.
\newblock \href {https://aclanthology.org/2020.lrec-1.6} {An annotated dataset of coreference in {E}nglish literature}.
\newblock In \emph{Proceedings of the Twelfth Language Resources and Evaluation Conference}, pages 44--54, Marseille, France. European Language Resources Association.

\bibitem[{Bamman et~al.(2019)Bamman, Popat, and Shen}]{bamman-etal-2019-annotated}
David Bamman, Sejal Popat, and Sheng Shen. 2019.
\newblock \href {https://doi.org/10.18653/v1/N19-1220} {An annotated dataset of literary entities}.
\newblock In \emph{Proceedings of the 2019 Conference of the North {A}merican Chapter of the Association for Computational Linguistics: Human Language Technologies, Volume 1 (Long and Short Papers)}, pages 2138--2144, Minneapolis, Minnesota. Association for Computational Linguistics.

\bibitem[{Bamman et~al.(2014)Bamman, Underwood, and Smith}]{Bamman2014}
David Bamman, Ted Underwood, and Noah~A. Smith. 2014.
\newblock \href {https://doi.org/10.3115/v1/P14-1035} {A {B}ayesian mixed effects model of literary character}.
\newblock In \emph{Proceedings of the 52nd Annual Meeting of the Association for Computational Linguistics (Volume 1: Long Papers)}, pages 370--379, Baltimore, Maryland. Association for Computational Linguistics.

\bibitem[{Baumgartner et~al.(2020)Baumgartner, Zannettou, Keegan, Squire, and Blackburn}]{Baumgartner2020}
Jason Baumgartner, Savvas Zannettou, Brian Keegan, Megan Squire, and Jeremy Blackburn. 2020.
\newblock \href {https://doi.org/10.1609/icwsm.v14i1.7347} {The pushshift reddit dataset}.
\newblock \emph{Proceedings of the International AAAI Conference on Web and Social Media}, 14(1):830--839.

\bibitem[{Brahman et~al.(2021)Brahman, Huang, Tafjord, Zhao, Sachan, and Chaturvedi}]{Brahman2021}
Faeze Brahman, Meng Huang, Oyvind Tafjord, Chao Zhao, Mrinmaya Sachan, and Snigdha Chaturvedi. 2021.
\newblock \href {https://doi.org/10.18653/v1/2021.findings-emnlp.150} {{``}let your characters tell their story{''}: A dataset for character-centric narrative understanding}.
\newblock In \emph{Findings of the Association for Computational Linguistics: EMNLP 2021}, pages 1734--1752, Punta Cana, Dominican Republic. Association for Computational Linguistics.

\bibitem[{Brooke et~al.(2015)Brooke, Hammond, and Hirst}]{brooke-etal-2015-gutentag}
Julian Brooke, Adam Hammond, and Graeme Hirst. 2015.
\newblock \href {https://doi.org/10.3115/v1/W15-0705} {{G}uten{T}ag: an {NLP}-driven tool for digital humanities research in the {P}roject {G}utenberg corpus}.
\newblock In \emph{Proceedings of the Fourth Workshop on Computational Linguistics for Literature}, pages 42--47, Denver, Colorado, USA. Association for Computational Linguistics.

\bibitem[{Cuesta-Lazaro et~al.(2022)Cuesta-Lazaro, Prasad, and Wood}]{CuestaLazaro2022}
Carolina Cuesta-Lazaro, Animesh Prasad, and Trevor Wood. 2022.
\newblock \href {https://doi.org/10.18653/v1/2022.acl-long.400} {What does the sea say to the shore? a {BERT} based {DST} style approach for speaker to dialogue attribution in novels}.
\newblock In \emph{Proceedings of the 60th Annual Meeting of the Association for Computational Linguistics (Volume 1: Long Papers)}, pages 5820--5829, Dublin, Ireland. Association for Computational Linguistics.

\bibitem[{Culpeper(1996)}]{Culpeper1996}
Jonathan Culpeper. 1996.
\newblock \href {https://doi.org/10.1016/0304-422X(95)00005-5} {Inferring character from texts: Attribution theory and foregrounding theory}.
\newblock \emph{Poetics}, 23(5):335--361.

\bibitem[{Devlin et~al.(2019)Devlin, Chang, Lee, and Toutanova}]{devlin-etal-2019-bert}
Jacob Devlin, Ming-Wei Chang, Kenton Lee, and Kristina Toutanova. 2019.
\newblock \href {https://doi.org/10.18653/v1/N19-1423} {{BERT}: Pre-training of deep bidirectional transformers for language understanding}.
\newblock In \emph{Proceedings of the 2019 Conference of the North {A}merican Chapter of the Association for Computational Linguistics: Human Language Technologies, Volume 1 (Long and Short Papers)}, pages 4171--4186, Minneapolis, Minnesota. Association for Computational Linguistics.

\bibitem[{Dinu and Uban(2017)}]{Dinu2017}
Liviu~P. Dinu and Ana~Sabina Uban. 2017.
\newblock \href {https://doi.org/10.18653/v1/W17-2210} {Finding a character{'}s voice: Stylome classification on literary characters}.
\newblock In \emph{Proceedings of the Joint {SIGHUM} Workshop on Computational Linguistics for Cultural Heritage, Social Sciences, Humanities and Literature}, pages 78--82, Vancouver, Canada. Association for Computational Linguistics.

\bibitem[{Elson et~al.(2010)Elson, Dames, and McKeown}]{Elson2010}
David Elson, Nicholas Dames, and Kathleen McKeown. 2010.
\newblock \href {https://aclanthology.org/P10-1015} {Extracting social networks from literary fiction}.
\newblock In \emph{Proceedings of the 48th Annual Meeting of the Association for Computational Linguistics}, pages 138--147, Uppsala, Sweden. Association for Computational Linguistics.

\bibitem[{Elson and McKeown(2010)}]{Elson2010a}
David Elson and Kathleen McKeown. 2010.
\newblock \href {https://doi.org/10.1609/aaai.v24i1.7720} {Automatic attribution of quoted speech in literary narrative}.
\newblock \emph{Proceedings of the AAAI Conference on Artificial Intelligence}, 24(1):1013--1019.

\bibitem[{Fischer et~al.(2019)Fischer, Börner, Göbel, Hechtl, Kittel, Milling, and Trilcke}]{Fischer2019}
Frank Fischer, Ingo Börner, Mathias Göbel, Angelika Hechtl, Christopher Kittel, Carsten Milling, and Peer Trilcke. 2019.
\newblock \href {https://doi.org/10.5281/zenodo.4284002} {Programmable {Corpora}: {Introducing} {DraCor}, an {Infrastructure} for the {Research} on {European} {Drama}}.
\newblock In \emph{Proceedings of {DH2019}: "{Complexities}", {Utrecht}, {July} 9–12, 2019}. Utrecht University.

\bibitem[{Gernsbacher et~al.(1992)Gernsbacher, Goldsmith, and Robertson}]{Gernsbacher1992}
Morton~Ann Gernsbacher, H~Hill Goldsmith, and Rachel~RW Robertson. 1992.
\newblock Do readers mentally represent characters' emotional states?
\newblock \emph{Cognition \& Emotion}, 6(2):89--111.

\bibitem[{He et~al.(2013)He, Barbosa, and Kondrak}]{He2013}
Hua He, Denilson Barbosa, and Grzegorz Kondrak. 2013.
\newblock \href {https://aclanthology.org/P13-1129} {Identification of speakers in novels}.
\newblock In \emph{Proceedings of the 51st Annual Meeting of the Association for Computational Linguistics (Volume 1: Long Papers)}, pages 1312--1320, Sofia, Bulgaria. Association for Computational Linguistics.

\bibitem[{He et~al.(2010)He, Kondrak, and Barbosa}]{He2010}
Hua He, Greg Kondrak, and Denilson Barbosa. 2010.
\newblock \href {https://api.semanticscholar.org/CorpusID:13246199} {The actor-topic model for extracting social networks in literary narrative}.

\bibitem[{Inoue et~al.(2022)Inoue, Pethe, Kim, and Skiena}]{Inoue2022}
Naoya Inoue, Charuta Pethe, Allen Kim, and Steven Skiena. 2022.
\newblock \href {https://doi.org/10.18653/v1/2022.findings-acl.81} {Learning and evaluating character representations in novels}.
\newblock In \emph{Findings of the Association for Computational Linguistics: ACL 2022}, pages 1008--1019, Dublin, Ireland. Association for Computational Linguistics.

\bibitem[{Joshi et~al.(2020)Joshi, Chen, Liu, Weld, Zettlemoyer, and Levy}]{Joshi2020}
Mandar Joshi, Danqi Chen, Yinhan Liu, Daniel~S. Weld, Luke Zettlemoyer, and Omer Levy. 2020.
\newblock \href {https://doi.org/10.1162/tacl_a_00300} {{SpanBERT: Improving Pre-training by Representing and Predicting Spans}}.
\newblock \emph{Transactions of the Association for Computational Linguistics}, 8:64--77.

\bibitem[{Khosla et~al.(2020)Khosla, Teterwak, Wang, Sarna, Tian, Isola, Maschinot, Liu, and Krishnan}]{Khosla2020}
Prannay Khosla, Piotr Teterwak, Chen Wang, Aaron Sarna, Yonglong Tian, Phillip Isola, Aaron Maschinot, Ce~Liu, and Dilip Krishnan. 2020.
\newblock \href {https://proceedings.neurips.cc/paper_files/paper/2020/file/d89a66c7c80a29b1bdbab0f2a1a94af8-Paper.pdf} {Supervised contrastive learning}.
\newblock In \emph{Advances in Neural Information Processing Systems}, volume~33, pages 18661--18673. Curran Associates, Inc.

\bibitem[{Labatut and Bost(2019)}]{Labatut2019}
Vincent Labatut and Xavier Bost. 2019.
\newblock \href {https://doi.org/10.1145/3344548} {Extraction and analysis of fictional character networks: A survey}.
\newblock \emph{ACM Computing Surveys}, 52(5):1–40.

\bibitem[{Li et~al.(2023)Li, Zhang, Li, and Yang}]{Li2023}
Dawei Li, Hengyuan Zhang, Yanran Li, and Shiping Yang. 2023.
\newblock \href {https://arxiv.org/abs/2310.13231} {Multi-level contrastive learning for script-based character understanding}.
\newblock \emph{Preprint}, arXiv:2310.13231.

\bibitem[{Michel et~al.(2024)Michel, Epure, Hennequin, and Cerisara}]{Michel2024}
Gaspard Michel, Elena Epure, Romain Hennequin, and Christophe Cerisara. 2024.
\newblock \href {https://aclanthology.org/2024.latechclfl-1.15} {Distinguishing fictional voices: a study of authorship verification models for quotation attribution}.
\newblock In \emph{Proceedings of the 8th Joint SIGHUM Workshop on Computational Linguistics for Cultural Heritage, Social Sciences, Humanities and Literature (LaTeCH-CLfL 2024)}, pages 160--171, St. Julians, Malta. Association for Computational Linguistics.

\bibitem[{Muzny et~al.(2017{\natexlab{a}})Muzny, Algee-Hewitt, and Jurafsky}]{Muzny2017a}
Grace Muzny, Mark Algee-Hewitt, and Dan Jurafsky. 2017{\natexlab{a}}.
\newblock \href {https://doi.org/10.1093/llc/fqx031} {{Dialogism in the novel: A computational model of the dialogic nature of narration and quotations}}.
\newblock \emph{Digital Scholarship in the Humanities}, 32:ii31--ii52.

\bibitem[{Muzny et~al.(2017{\natexlab{b}})Muzny, Fang, Chang, and Jurafsky}]{Muzny2017}
Grace Muzny, Michael Fang, Angel Chang, and Dan Jurafsky. 2017{\natexlab{b}}.
\newblock \href {https://aclanthology.org/E17-1044} {A two-stage sieve approach for quote attribution}.
\newblock In \emph{Proceedings of the 15th Conference of the {E}uropean Chapter of the Association for Computational Linguistics: Volume 1, Long Papers}, pages 460--470, Valencia, Spain. Association for Computational Linguistics.

\bibitem[{O{'}Keefe et~al.(2012)O{'}Keefe, Pareti, Curran, Koprinska, and Honnibal}]{okeefe-etal-2012-sequence}
Timothy O{'}Keefe, Silvia Pareti, James~R. Curran, Irena Koprinska, and Matthew Honnibal. 2012.
\newblock \href {https://aclanthology.org/D12-1072} {A sequence labelling approach to quote attribution}.
\newblock In \emph{Proceedings of the 2012 Joint Conference on Empirical Methods in Natural Language Processing and Computational Natural Language Learning}, pages 790--799, Jeju Island, Korea. Association for Computational Linguistics.

\bibitem[{Reimers and Gurevych(2019)}]{reimers-gurevych-2019-sentence}
Nils Reimers and Iryna Gurevych. 2019.
\newblock \href {https://doi.org/10.18653/v1/D19-1410} {Sentence-{BERT}: Sentence embeddings using {S}iamese {BERT}-networks}.
\newblock In \emph{Proceedings of the 2019 Conference on Empirical Methods in Natural Language Processing and the 9th International Joint Conference on Natural Language Processing (EMNLP-IJCNLP)}, pages 3982--3992, Hong Kong, China. Association for Computational Linguistics.

\bibitem[{Rivera-Soto et~al.(2021)Rivera-Soto, Miano, Ordonez, Chen, Khan, Bishop, and Andrews}]{RiveraSoto2021}
Rafael~A. Rivera-Soto, Olivia~Elizabeth Miano, Juanita Ordonez, Barry~Y. Chen, Aleem Khan, Marcus Bishop, and Nicholas Andrews. 2021.
\newblock \href {https://doi.org/10.18653/v1/2021.emnlp-main.70} {Learning universal authorship representations}.
\newblock In \emph{Proceedings of the 2021 Conference on Empirical Methods in Natural Language Processing}, pages 913--919, Online and Punta Cana, Dominican Republic. Association for Computational Linguistics.

\bibitem[{Sang et~al.(2022)Sang, Mou, Yu, Yao, Li, and Stanton}]{Sang2022}
Yisi Sang, Xiangyang Mou, Mo~Yu, Shunyu Yao, Jing Li, and Jeffrey Stanton. 2022.
\newblock \href {https://doi.org/10.18653/v1/2022.naacl-main.317} {{TVS}how{G}uess: Character comprehension in stories as speaker guessing}.
\newblock In \emph{Proceedings of the 2022 Conference of the North American Chapter of the Association for Computational Linguistics: Human Language Technologies}, pages 4267--4287, Seattle, United States. Association for Computational Linguistics.

\bibitem[{Sims and Bamman(2020)}]{Sims2020}
Matthew Sims and David Bamman. 2020.
\newblock \href {https://doi.org/10.18653/v1/2020.emnlp-main.47} {Measuring information propagation in literary social networks}.
\newblock In \emph{Proceedings of the 2020 Conference on Empirical Methods in Natural Language Processing (EMNLP)}, pages 642--652, Online. Association for Computational Linguistics.

\bibitem[{Su et~al.(2023)Su, Xu, Xu, Li, and Huangfu}]{Su2023}
Zhenlin Su, Liyan Xu, Jin Xu, Jiangnan Li, and Mingdu Huangfu. 2023.
\newblock \href {https://arxiv.org/abs/2312.14590} {Sig: Speaker identification in literature via prompt-based generation}.
\newblock \emph{Preprint}, arXiv:2312.14590.

\bibitem[{Vishnubhotla et~al.(2019)Vishnubhotla, Hammond, and Hirst}]{Vishnubhotla2019}
Krishnapriya Vishnubhotla, Adam Hammond, and Graeme Hirst. 2019.
\newblock \href {https://doi.org/10.18653/v1/W19-2504} {Are fictional voices distinguishable? classifying character voices in modern drama}.
\newblock In \emph{Proceedings of the 3rd Joint {SIGHUM} Workshop on Computational Linguistics for Cultural Heritage, Social Sciences, Humanities and Literature}, pages 29--34, Minneapolis, USA. Association for Computational Linguistics.

\bibitem[{Vishnubhotla et~al.(2022)Vishnubhotla, Hammond, and Hirst}]{Vishnubhotla2022}
Krishnapriya Vishnubhotla, Adam Hammond, and Graeme Hirst. 2022.
\newblock \href {https://aclanthology.org/2022.lrec-1.628} {The project dialogism novel corpus: A dataset for quotation attribution in literary texts}.
\newblock In \emph{Proceedings of the Thirteenth Language Resources and Evaluation Conference}, pages 5838--5848, Marseille, France. European Language Resources Association.

\bibitem[{Vishnubhotla et~al.(2023)Vishnubhotla, Rudzicz, Hirst, and Hammond}]{Vishnubhotla2023}
Krishnapriya Vishnubhotla, Frank Rudzicz, Graeme Hirst, and Adam Hammond. 2023.
\newblock \href {https://doi.org/10.18653/v1/2023.acl-short.64} {Improving automatic quotation attribution in literary novels}.
\newblock In \emph{Proceedings of the 61st Annual Meeting of the Association for Computational Linguistics (Volume 2: Short Papers)}, pages 737--746, Toronto, Canada. Association for Computational Linguistics.

\bibitem[{Wegmann et~al.(2022)Wegmann, Schraagen, and Nguyen}]{Wegmann2022}
Anna Wegmann, Marijn Schraagen, and Dong Nguyen. 2022.
\newblock \href {https://doi.org/10.18653/v1/2022.repl4nlp-1.26} {Same author or just same topic? towards content-independent style representations}.
\newblock In \emph{Proceedings of the 7th Workshop on Representation Learning for NLP}, pages 249--268, Dublin, Ireland. Association for Computational Linguistics.

\bibitem[{Zhang et~al.(2023)Zhang, Li, Zhang, Zhan, Wang, Zhou, and Qiu}]{zhang-etal-2023-speechgpt}
Dong Zhang, Shimin Li, Xin Zhang, Jun Zhan, Pengyu Wang, Yaqian Zhou, and Xipeng Qiu. 2023.
\newblock \href {https://doi.org/10.18653/v1/2023.findings-emnlp.1055} {{S}peech{GPT}: Empowering large language models with intrinsic cross-modal conversational abilities}.
\newblock In \emph{Findings of the Association for Computational Linguistics: EMNLP 2023}, pages 15757--15773, Singapore. Association for Computational Linguistics.

\bibitem[{Šeļa et~al.(2023)Šeļa, Nagy, Byszuk, Hernández-Lorenzo, Szemes, and Eder}]{Sela2023}
Artjoms Šeļa, Ben Nagy, Joanna Byszuk, Laura Hernández-Lorenzo, Botond Szemes, and Maciej Eder. 2023.
\newblock \href {https://arxiv.org/abs/2301.05659} {From stage to page: language independent bootstrap measures of distinctiveness in fictional speech}.
\newblock \emph{Preprint}, arXiv:2301.05659.

\end{thebibliography}

\appendix

\section{Details on the Construction and Evaluation of DramaCV}
\label{sec:appendix_a}
Each play collected with GutenTag is encoded in a `XML' file, containing various tags such as the \texttt{<speaker>} tag that indicates the speaker of a line.
We parse each XML file, removing plays written by non English-speaking authors, and attribute each line to its speaker using the \texttt{<speaker>} tag.

For the Scene split, we automatically check if a play contains a \texttt{<scene>} or \texttt{<act>} tag.
We segment plays by default with the \texttt{<scene>} tag, and use the \texttt{<act>} tag if it does not contain a scene tag.
We removed plays that did not contain any of these tags.
For the Play split, we kept all plays written by English-speaking authors.

In the Authorship Verification terminology, queries and targets are collections of documents written by either the same or different authors.
Given a query written by author A and multiple targets written by either the same or different authors, the objective of AV is to recognize all targets written by author A.
Details of queries and target construction and our evaluation setup are provided below:

\paragraph{Queries/Targets.} Given characters $(c_1,\dots,c_n)$ occurring in a single segment (a scene/act for the Scene split, or the entire play for the Play split), we randomly sample half of their utterances to construct queries $(q_{c_1},\dots,q_{c_n})$ and use the remaining half to construct targets $(t_{c_1},\dots,t_{c_n})$.
For the train split, we instead randomly sample 8 utterances to build queries and 8 different utterances to build targets.

\paragraph{DramaCV Evaluation.} Using any model $\mathcal{M}$ that is able to encode collections of documents in a feature vector, we encode each query and target independently. 
Then, we compare the cosine similarity of query vectors with targets: $similarity(\mathcal{M}(q), \mathcal{M}(t))$ $\forall q \in q_{c_1},\dots,q_{c_n}$ and $\forall t \in t_{c_1},\dots,t_{c_n}$.
For a query $q_{c}$ uttered by character $c$, we calculate an AUC score over the resulting cosine similarities with targets, where a label $1$ is assigned to $t_c$ and $0$ to other targets.
This AUC represents the ability of $\mathcal{M}$ to assign an higher cosine similarities between queries and targets written by the same author.
We average this AUC score over each query in a single segment to obtain an AUC score for this particular segment.

\paragraph{PDNC Evaluation.}  We use the evaluation setup of \citet{Michel2024} to evaluate our models on the characters of PDNC.
This setup proposes two types of evaluation: Character-Character (CC) and Character-Quotes (CQ).

The CC evaluation is similar to the one described above, but the queries are constructed using only the explicit quotes of a character in a single chapter, and targets are constructed using every other utterance in the other chapters.
The AUC computed scores how a model is able to assign a higher cosine-similarity between a collection of a few explicit quotes of character, and its other utterances in the novel.

In the CQ setup, queries are similar to CC, but a target is defined as a single utterance in the held-out chapters, without access to contextual information.
Therefore, the AUC computed is scoring how similar the collection of a few explicit quotes of character is to its other individual quotes.
This task is inherently harder than the CC setup, as an individual quote might be very different from other quotes (e.g ``Hey, how are you Lizzy?''). 

\section{Performance on all Test Plays}
\label{sec:appendix_b}
We report performance of $\text{UAR}_{Scene}$ and $\text{UAR}_{Play}$ on their respective test split in Table~\ref{tab:results_uar_scene} and Table~\ref{tab:results_uar_play}.

\section{Computing Information}
We used a 32-core Intel Xeon Gold 6244 CPU @ 3.60GHz CPU with 128GB RAM equipped with 3 RTX A5000 GPUs with 24GB RAM each and a single Nvidia A40 with 45GB RAM.
We trained $\text{UAR}_{Scene}$ and $\text{UAR}_{Play}$ with the Nvidia A40, requiring gradient-checkpointing to fit everything in the GPU RAM.
The training took around 1 hour for both models.
We use a single RTX A5000 for training $\text{BookNLP+}_{reimp}$ and its variant with character embeddings.
Training on a single cross-validation split took around 6 hours for each of the proposed methods.
Inference times are shorter, where quotation attribution on a full novel can be done in a few minutes only.

\begin{table*}[]
    \centering
    \small
    \begin{tabular}{llc}
        \toprule
         Play & Author & AUC\\
         \midrule
Justice & John Galsworthy &  95.6\\
The Alchemist & Ben Jonson &  91.8\\
Hamlet & William Shakespeare &  86.5\\
Plays : Third Series & John Galsworthy &  86.3\\
The works of John Dryden & John Dryden &  85.2\\
The Life of Timon of Athens & William Shakespeare &  84.2\\
King Henry the Fifth & William Shakespeare &  83.7\\
Night Must Fall : a Play in Three Acts & Emlyn Williams &  83.7\\
The Spanish Tragedy & Thomas Kyd &  83.2\\
The works of John Dryden & John Dryden &  83.0\\
The Scarlet Stigma: A Drama in Four Acts & Nathaniel Hawthorne &  82.4\\
King John & William Shakespeare &  81.8\\
The Works of Charles and Mary Lamb & Mary Lamb &  80.8\\
The Scornful Lady & John Fletcher &  80.8\\
Married Life A Comedy, in Three Acts & John Baldwin Buckstone &  80.0\\
The Flutter of the Goldleaf; and Other Plays & Frederick Peterson &  77.5\\
Single Life	A Comedy, in Three Acts & John Baldwin Buckstone &  75.9\\
The Countess Cathleen & W. B. (William Butler) Yeats &  75.6\\
De Turkey and De Law: A Comedy in Three Acts & Zora Neale Hurston &  66.4\\
    \bottomrule
    \end{tabular}
    \caption{AUC per novel on the Scene test split of DramaCV.}
    \label{tab:results_uar_scene}
\end{table*}

\begin{table*}[]
    \centering
    \small
    \begin{tabular}{llc}
        \toprule
         Play & Author & AUC\\
         \midrule
Bride Roses & William Dean Howells &  100.0\\
Fanny's First Play & Bernard Shaw &  100.0\\
Five Little Plays & Alfred Sutro &  98.6\\
John Bull's Other Island & Bernard Shaw &  95.4\\
Mrs. Dot: A Farce & William Somerset Maugham &  94.5\\
The Earl of Essex: A Tragedy, in Five Acts & Henry Jones &  94.4\\
The Way of the World & William Congreve &  92.7\\
Plays : Fifth Series & John Galsworthy &  92.5\\
Representative Plays by American Dramatists: 1856-1911: The New York Idea & Langdon Elwyn Mitchell &  91.9\\
Plays: Lady Frederick, The Explorer, A Man of Honour & William Somerset Maugham &  90.3\\
Great Catherine & Bernard Shaw &  89.3\\
Hypolympia	Or, The Gods in the Island, an Ironic Fantasy & Edmund Gosse &  87.1\\
The works of John Dryden & John Dryden &  87.0\\
The Grecian Daughter & Arthur Murphy &  85.5\\
Dramatic Technique & George Pierce Baker &  81.6\\
Lyre and Lancet: A Story in Scenes & F. Anstey &  80.2\\
Beaumont and Fletcher's Works, Vol. 5 & Francis Beaumont &  79.0\\
King Henry VI, Part 3 & William Shakespeare &  78.1\\
Plays and Lyrics & Cale Young Rice &  78.0\\
The Easiest Way	Representative Plays by American Dramatists: 1856-1911 & Eugene Walter &  77.6\\
Charles Di Tocca: A Tragedy & Cale Young Rice &  77.3\\
St. Patrick's day, or, the scheming lieutenant : a farce in one act & Richard Brinsley Sheridan &  76.7\\
Deirdre of the Sorrows & John Millington Synge &  76.3\\
Ambrose Gwinett	or, a sea-side story : a melo-drama, in three acts & Douglas William Jerrold &  75.8\\
Henry V & William Shakespeare &  75.3\\
Box and Cox: A Romance of Real Life in One Act. & John Maddison Morton &  75.0\\
The Recruiting Officer & George Farquhar &  74.7\\
Yolanda of Cyprus & Cale Young Rice &  74.3\\
    \bottomrule
    \end{tabular}
    \caption{AUC per novel on the Play test split of DramaCV.}
    \label{tab:results_uar_play}
\end{table*}

\end{document}